\documentclass[letterpaper,10pt,conference]{ieeeconf}

\usepackage{graphicx}
\usepackage{amsmath,amssymb,amsfonts}
\usepackage{booktabs}
\usepackage{multirow}
\usepackage{threeparttable}
\usepackage{hyperref}
\usepackage{censor}
\usepackage{stfloats}

\IEEEoverridecommandlockouts
\title{\LARGE \bf
Execution-Aware Pre-Execution Ranking for Grasp-Conditioned Robotic Placement
}

\author{%
  Tianyuan Liu,
  Rutherford Agbeshi Patamia,
  Benjamin Champion,
  Richard Dazeley,
  Akansel Cosgun$^*$%
  \thanks{$^*$All authors are with Deakin University, Australia.
  Email: \{tianyuan.liu, r.patamia, richard.dazeley, benjamin.champion, akan.cosgun\}@deakin.edu.au}
}

\makeatletter
\let\origthebibliography\thebibliography
\renewcommand{\thebibliography}[1]{%
  \origthebibliography{#1}%
  \fontsize{7.5}{8.0}\selectfont
  \setlength{\itemsep}{-0.8pt}%
}
\makeatother

\begin{document}

\maketitle
\thispagestyle{empty}
\pagestyle{empty}

\begin{abstract}
A geometrically valid placement can still be difficult to execute because the selected grasp changes the required end-effector pose, collision geometry, and transport motion. Placement is formulated as a pre-execution ranking problem in which supplied grasp--placement candidates are scored before planning. The model combines a typed target-conditioned point cloud with three pose descriptors and hierarchical heads for planning success and execution success conditioned on planning. On a 30-object, 1,235-scene dataset with scene-group-held-out splits, three-seed top-1 success on covered test groups reaches $85.63\pm1.08\%$ for joint selection and $79.84\pm0.16\%$ for fixed-target ranking. For the designated frozen seed-42 checkpoint, top-1 success improves from 72.84\% to 85.78\% over full-pool cuMotion for joint ranking and from 59.65\% to 79.67\% for fixed-target ranking. Frozen transfer to xArm7/MoveIt requires no xArm-specific retraining. Across 27 locked cases, 13 complete end to end (48.15\%). Of the 16 cases that pass Top-5 preflight and begin execution, 13 succeed (81.25\%). Candidate-level deployment-feasibility prediction reaches 81.25\% recall, 85.20\% specificity, and 83.23\% balanced accuracy.
\end{abstract}

\section{Introduction}

\begin{figure*}[!b]
    \centering
    \includegraphics[width=\textwidth]{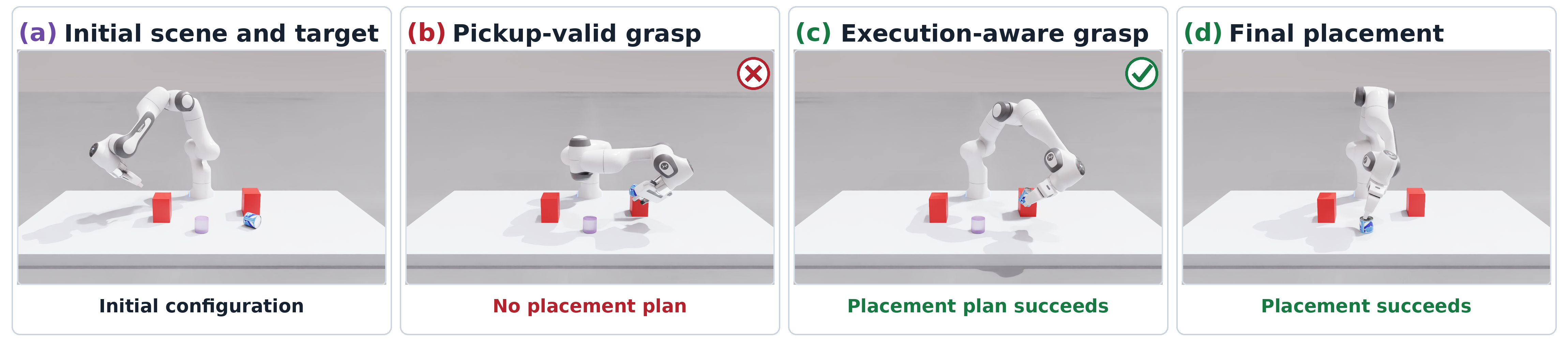}
    \caption{Grasp choice determines placement feasibility. For the same initial scene and target pose, both candidate grasps pass the pickup-and-lift check. The pickup-valid grasp in (b) admits no complete placement plan, while the execution-aware ranker selects the grasp in (c), which supports a valid placement motion and produces the successful final state in (d).}
    \label{fig:moneyshot}
\end{figure*}

A placement pose that is geometrically valid for an object is not necessarily a good action for the robot. Once a grasp is chosen, the object pose relative to the gripper is fixed. That choice changes the end-effector pose required at the destination, the collision geometry, and the motion needed to carry the object there. As a result, two grasps for the same target can have very different outcomes even when the target itself is unchanged, as illustrated in Fig.~\ref{fig:moneyshot}.

A common way to handle a large candidate set is \emph{rank-then-plan}: score the grasp--placement pairs cheaply, then spend motion-planning effort on the most promising ones. This efficiency is especially important in long-horizon manipulation, where repeated planning attempts at poor candidates accumulate across many sequential decisions and can dominate overall task time. Prior work has addressed geometric placement~\cite{extra,r2}, learned placement from partial observations~\cite{r3,r4}, coupled grasp--placement planning~\cite{r5,r6,r7,r8}, and learned feasibility guidance for manipulation planning~\cite{r9,r10,r11}. The remaining issue is what the ranking score should represent. Endpoint IK and local collision checks are useful, but they do not tell us whether a complete path can be planned or whether the object will remain valid through the whole task trajectory. Running a full planner over every candidate provides more information, but its cost grows quickly with the pool size.

This paper focuses on \emph{execution-aware pre-execution ranking}. The model receives no planner result, trajectory, measured post-lift state, or execution outcome at inference time. It uses only a target-conditioned typed scene and candidate-defined pose quantities that are known before planning. From these inputs, it estimates (i) whether the requested placement motion can be planned and (ii) whether the remaining execution succeeds given planning success. Their product gives the candidate score.

\begin{figure*}[t]
    \centering
    \includegraphics[width=0.86\textwidth]{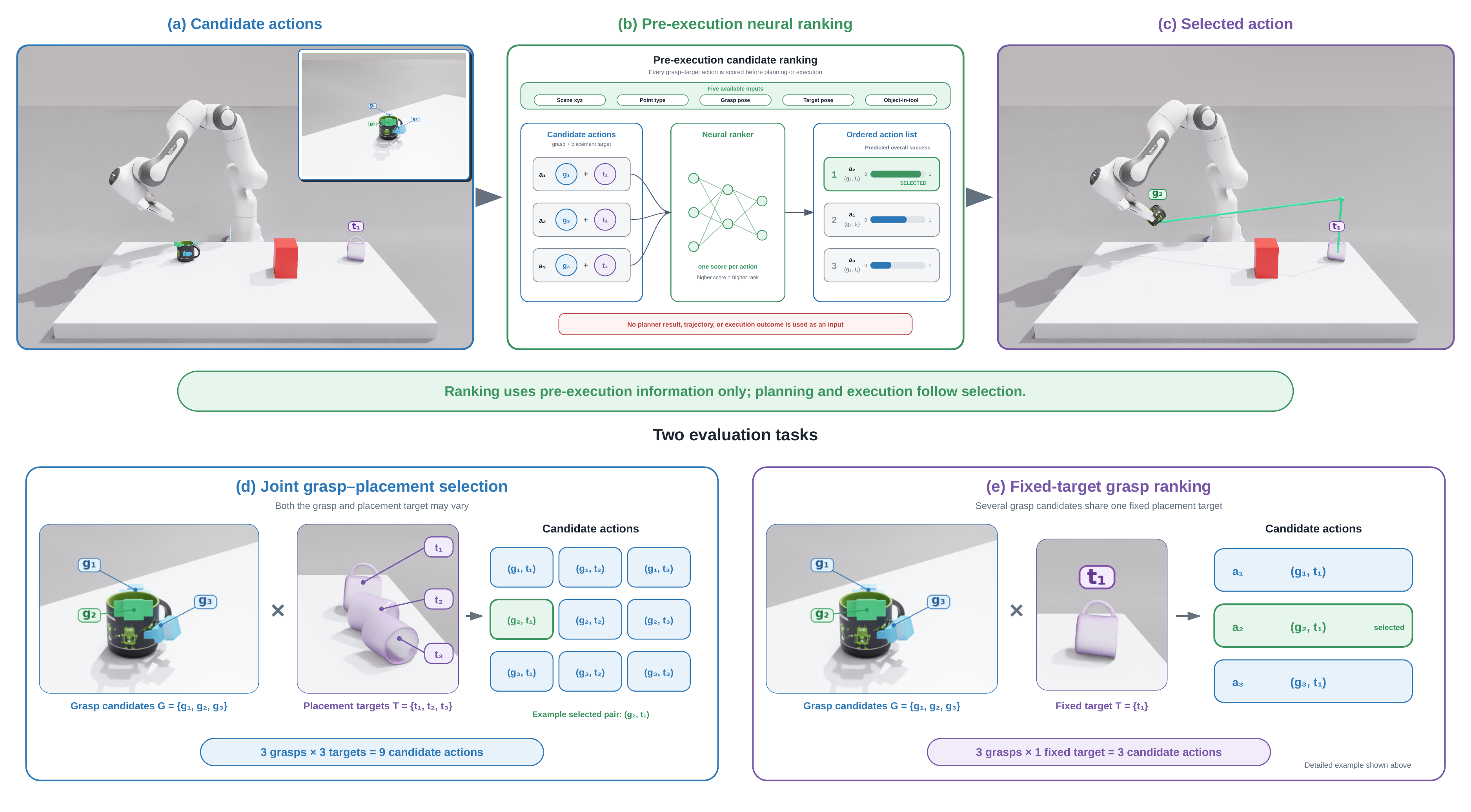}
    \caption{Problem overview. Supplied grasps are paired with placement targets and ranked using only pre-execution information. Joint selection ranks all grasp--target pairs. Fixed-target ranking compares grasps for one specified destination. Planning and execution occur after ranking.}
    \label{fig:overview}
\end{figure*}

Two ranking tasks are studied, as shown in Fig.~\ref{fig:overview}. In \emph{joint grasp--placement selection}, all supplied grasp--target pairs are ranked together. In \emph{fixed-target grasp ranking}, the destination is specified and only grasps for that target are ranked. Candidate-pool coverage---whether the supplied pool contains any successful action---is reported separately from ranking accuracy within covered groups. Frozen-model transfer to xArm7/MoveIt is also evaluated without xArm-specific retraining.

The main contributions are:
\begin{itemize}
    \item an execution-aware ranking formulation with a strict five-input inference contract and hierarchical planning/execution supervision.
    \item a 30-object, 1,235-scene execution-grounded benchmark with scene-group-held-out splits and separate joint and fixed-target evaluation, and
    \item a full-pool comparison with endpoint checks and cuMotion, together with three-seed evaluation, timing, candidate-pool solvability analysis, label-blind fresh-execution validation, and frozen-model cross-embodiment transfer to an xArm7/MoveIt system without xArm-specific retraining.
\end{itemize}
Unlike the earlier endpoint-feasibility model in~\cite{r9}, which learned IK and collision proxies, the present ranker is supervised by full motion-planning outcomes and conditional downstream execution outcomes.

\section{Related Work}

\subsection{Placement and Grasp--Place Coupling}

Robotic placement combines stability, geometric compatibility, reachability, and collision-free motion. Model-based methods generate or search for feasible placements using support geometry, stability criteria, and robot constraints~\cite{extra,r2}. Learning-based methods relax the need for complete object models, for example by predicting stable placements from partial point clouds~\cite{r3} or using tactile feedback to correct the object before release~\cite{r4}. These methods mainly answer where or how an object should be placed.

The grasp cannot be considered independently of that decision. It fixes the object relative to the end effector, which changes the target robot pose and the transport geometry. Prior work has optimized grasp and placement jointly with learned grasp models and geometric collision objectives~\cite{r5}, searched jointly over grasps, placements, and motions~\cite{r6}, and incorporated grasp--placement compatibility into regrasp or packing pipelines~\cite{r7,r8}. The present setting keeps this coupled action representation but focuses on ranking a supplied pool before expensive planning.

\subsection{Learning to Guide Planning}

Learned feasibility models can reduce expensive geometric queries in task-and-motion planning~\cite{r10}. More recent work predicts action feasibility directly from 3D environments~\cite{aitbouhsain2023action} and jointly predicts action and grasp feasibility~\cite{aitbouhsain2023multitask}. Learned geometric functionals have also been used inside manipulation-planning objectives~\cite{r11}. Earlier placement work~\cite{r9} ranked grasp-dependent candidates from inexpensive endpoint IK and collision signals. The present work instead learns from full planning outcomes and later physics-based execution outcomes. A collision-free endpoint may still fail along the path or after release.

\subsection{Geometric Representation and Simulation Supervision}

Synthetic manipulation supervision ranges from analytic grasp labels in Dex-Net~\cite{mahler2018dex} to modern simulators and large 3D asset collections~\cite{isaacsim,deitke2023objaverse}. Point-based manipulation~\cite{kuang2026dex4d}, implicit pose encodings~\cite{moreau2023imposing}, continuous rotation representations~\cite{r13}, and Fourier features~\cite{r14} provide useful representation choices. The model uses a typed target-conditioned scene with compact pose encoders; the contribution is the execution-aware ranking formulation and its connection between pre-execution inputs and planning/execution supervision, rather than a new point-cloud backbone.

\section{Problem Formulation}
\label{sec:problem_formulation}

For a scene $\mathcal{S}$, let $\mathcal{G}=\{g_i\}_{i=1}^{N_g}$ denote the supplied grasp candidates and $\mathcal{P}=\{p_j\}_{j=1}^{N_p}$ the supplied placement targets. A candidate action is
\begin{equation}
    a_{ij}=(g_i,p_j).
    \label{eq:candidate_action}
\end{equation}
The method does not generate grasps or targets. It only orders these supplied actions.

\paragraph{Joint grasp--placement selection}
When several placement targets are acceptable, all supplied grasp--target pairs are considered together:
\begin{equation}
\mathcal{A}_{\mathrm{joint}}
=
\{(g_i,p_j):g_i\in\mathcal{G},\,p_j\in\mathcal{P}\}.
\label{eq:joint_action_set}
\end{equation}
Both the grasp and placement target may vary, so the ranker selects a complete grasp--placement action.

\paragraph{Fixed-target grasp ranking}
When the destination is specified externally as $p^*$, the candidate set becomes
\begin{equation}
\mathcal{A}_{\mathrm{fixed}}(p^*)
=
\{(g_i,p^*)\}_{i=1}^{N_g}.
\label{eq:fixed_action_set}
\end{equation}
All candidates now share the same placement target, so only the grasp changes. This setting directly tests which validated grasp is most compatible with one required destination.

For each candidate, the ranker uses only information available before motion planning or execution and predicts
\begin{equation}
    r_{ij}=f_\theta(\mathcal{S},a_{ij}).
    \label{eq:ranking_function}
\end{equation}
Here, $r_{ij}$ is a scalar pre-execution ranking score for action $a_{ij}=(g_i,p_j)$. A larger value means the action is predicted to be more likely to complete the placement pipeline successfully. Candidate actions are ordered from highest to lowest $r_{ij}$, and the downstream motion planner evaluates them in that order. The ranker is a ranking step, not a replacement for motion planning. By moving promising actions earlier in the list, it reduces unnecessary planner calls. This saving adds up across repeated decisions in long-horizon tasks.

\section{Execution-Aware Ranking Method}

\subsection{Scene and Candidate Representation}

\begin{figure}[t]
    \centering
    \includegraphics[width=\columnwidth]
    {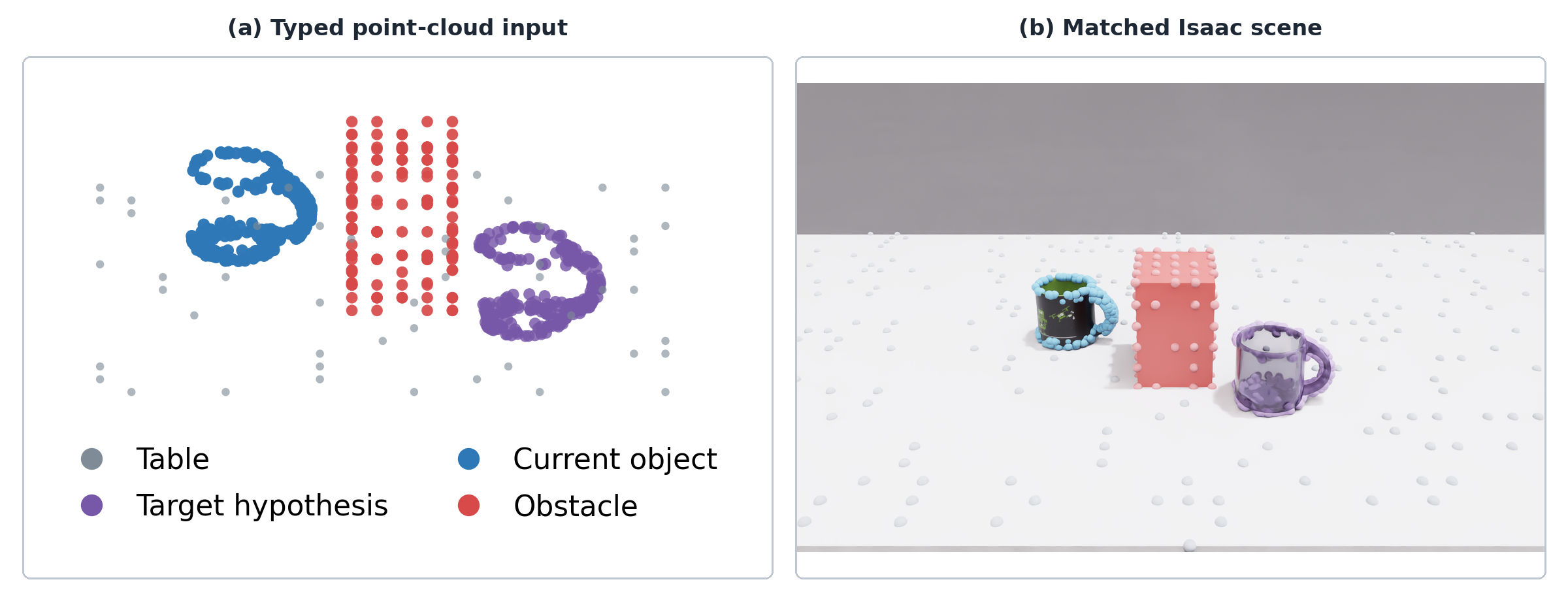}
    \caption{Target-conditioned point cloud. (a) Typed input with the current object, virtual target hypothesis, support surface, and obstacle. (b) Matched Isaac scene; the target hypothesis is virtual, not a second observed object.}
    \label{fig:pcd_scene}
\end{figure}

For placement target $p_j$, the scene point cloud in the robot
base frame contains the current object, support surface, and
obstacles. This observed geometry is retained and augmented with a virtual
copy of the manipulated-object points transformed to the
requested target pose. The resulting target-conditioned scene is
\begin{equation}
\mathbf{X}_j =
\mathbf{X}_{\mathrm{env}}
\cup
\mathbf{X}^{\mathrm{cur}}_{\mathrm{obj}}
\cup
\mathbf{X}^{\mathrm{tar},j}_{\mathrm{obj}}
\cup
\mathbf{X}_{\mathrm{obs}},
\label{eq:target_conditioned_scene}
\end{equation}
where the four point types denote static environment, current
object, the virtual target-object hypothesis, and obstacle.
The virtual target copy makes the intended object occupancy explicit
without replacing the observed current-object geometry
(Fig.~\ref{fig:pcd_scene}).

For candidate action $a_{ij}=(g_i,p_j)$, the complete model
input is
\begin{equation}
\mathbf{x}_{ij}
=
\left(
\mathbf{X}_j,\,
\mathbf{c}_j,\,
g_i,\,
p_j,\,
{}^{E}T_{O,i}
\right),
\label{eq:model_input}
\end{equation}
where $\mathbf{X}_j$ is the target-conditioned point cloud,
$\mathbf{c}_j$ contains the corresponding point-type IDs,
$g_i$ is the candidate grasp pose in the robot base frame,
$p_j$ is the requested target pose, and
${}^{E}T_{O,i}$ describes the candidate-defined object pose
relative to the initial grasp frame.

The three pose inputs are each represented by a 3-D translation
and a continuous 6-D rotation representation. All quantities in
Eq.~\eqref{eq:model_input} are defined before motion planning
or execution. In particular, the model receives no measured
post-lift robot state, measured post-lift object-relative pose,
planner status, trajectory, or physics outcome. These quantities
are reserved for supervision and evaluation.

\subsection{Encoding and Hierarchical Prediction}

A PointNet++ backbone~\cite{pointnetpp2017} encodes $(\mathbf{X}_j,\mathbf{c}_j)$
into a 512-dimensional scene feature. Point types use learned
categorical embeddings rather than a continuous scalar channel.
Three independent pose branches encode the grasp pose,
object-relative pose, and target pose. Each branch uses the
3-D translation and continuous 6-D rotation representation,
followed by Fourier features and multilayer perceptrons.

\begin{figure*}[t]
    \centering
    \includegraphics[
        width=0.85\textwidth,
        trim={10 15 5 0},
        clip
    ]{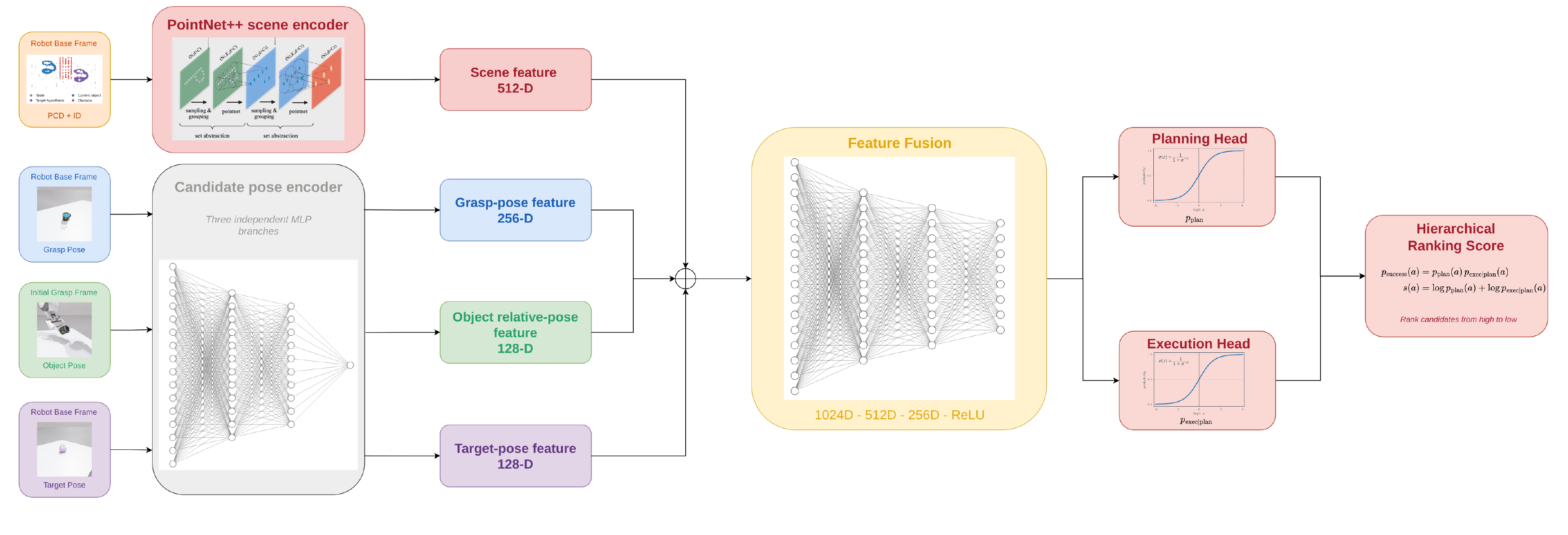}
    \caption{Model architecture. PointNet++ encodes the
    target-conditioned typed scene, while three independent
    pose branches encode the grasp pose, object-relative pose,
    and target pose. The fused representation feeds separate
    planning and execution heads whose probabilities form the
    final ranking score.}
    \label{fig:model_architecture}
\end{figure*}

The grasp-pose branch produces a 256-D feature, while the
object-relative-pose and target-pose branches each produce
128-D features. The four features are concatenated into a
1024-D representation and passed through shared 512-D and
256-D fusion layers before reaching two prediction heads, as
shown in Fig.~\ref{fig:model_architecture}.

\paragraph{Placement-planning feasibility.}
The planning head estimates
\begin{equation}
p_{\mathrm{plan}}^{ij}
=
P\!\left(
y_{\mathrm{plan}}^{ij}=1
\mid
\mathbf{x}_{ij}
\right),
\label{eq:planning_probability}
\end{equation}
where $y_{\mathrm{plan}}^{ij}=1$ indicates that a valid
placement motion can be planned for candidate action
$a_{ij}$.

\paragraph{Placement completion given a valid plan.}
For candidates that reach the planning-success stage, the
execution head estimates
\begin{equation}
p_{\mathrm{exec|plan}}^{ij}
=
P\!\left(
y_{\mathrm{pipe}}^{ij}=1
\mid
y_{\mathrm{plan}}^{ij}=1,\,
\mathbf{x}_{ij}
\right),
\label{eq:execution_probability}
\end{equation}
where $y_{\mathrm{pipe}}^{ij}=1$ denotes successful transport,
placement, release, settling, and final-pose validation after
a valid plan has been obtained.

The corresponding full-pipeline success probability is
\begin{equation}
p_{\mathrm{succ}}^{ij}
=
p_{\mathrm{plan}}^{ij}
p_{\mathrm{exec|plan}}^{ij}.
\label{eq:overall_probability}
\end{equation}

For numerical stability, candidate actions are ranked using the
equivalent log-domain score
\begin{equation}
r_{ij}
=
\log p_{\mathrm{plan}}^{ij}
+
\log p_{\mathrm{exec|plan}}^{ij}.
\label{eq:ranking_score}
\end{equation}
Here, $p_{\mathrm{succ}}^{ij}$ is the predicted probability that
candidate action $a_{ij}$ completes the full placement pipeline,
while $r_{ij}=\log p_{\mathrm{succ}}^{ij}$ is the scalar score
used to order candidates. Since the logarithm is monotonic,
sorting candidates by $r_{ij}$ produces exactly the same
ordering as sorting them by $p_{\mathrm{succ}}^{ij}$.

\subsection{Supervision}

The planning label supervises the planning head for every
candidate. The execution head is supervised only for candidates
that successfully pass planning, because an execution outcome
is not defined for an action that never reaches that stage.
This masking avoids assigning artificial execution failures to
planning failures. The final objective is $\mathcal{L}=2\mathcal{L}_{\mathrm{plan}}+\mathcal{L}_{\mathrm{exec}}+0.1\mathcal{L}_{\mathrm{rank}}$, where $\mathcal{L}_{\mathrm{plan}}$ is binary cross-entropy over all candidates, $\mathcal{L}_{\mathrm{exec}}$ is binary cross-entropy masked to $y_{\mathrm{plan}}=1$, and $\mathcal{L}_{\mathrm{rank}}$ is a listwise ranking loss for groups with valid ranking supervision.

Both labels are generated by the planning and physics pipeline
described in Sec.~V and are used only for training and
evaluation. At test time, Eq.~\eqref{eq:ranking_score} is
computed entirely from the five pre-execution inputs in
Eq.~\eqref{eq:model_input}, before the downstream motion
planner is called.

\section{Dataset and Experimental Setup}
\label{sec:dataset_experiments}

\subsection{Execution-Grounded Dataset and Splits}

The dataset is constructed in Isaac Sim~\cite{isaacsim} with a Franka Emika Panda and cuMotion planning, using 30 known objects spanning cans, mugs, and cuboid objects. It contains 1,235 valid scenes with clean, one-obstacle, and two-obstacle environments. The final split contains 153,072 training candidates, 48,200 validation candidates, and 48,376 candidates in the untouched test set.

Each scene specifies an initial object configuration, a set of supplied grasp candidates, and placement targets. The grasps have already passed the upstream grasp-admission and pick--lift checks. The learned model begins at the placement-ranking stage. Each retained grasp is paired with the available targets to form the candidate actions. For every action, the pipeline first evaluates motion planning. Candidates with a valid plan are then executed in physics through transport, placement, release, settling, and final-pose validation. These outcomes provide $y_{\mathrm{plan}}^{ij}$ and $y_{\mathrm{pipe}}^{ij}$.

\begin{figure*}[!t]
    \centering
    \includegraphics[width=0.8\textwidth]{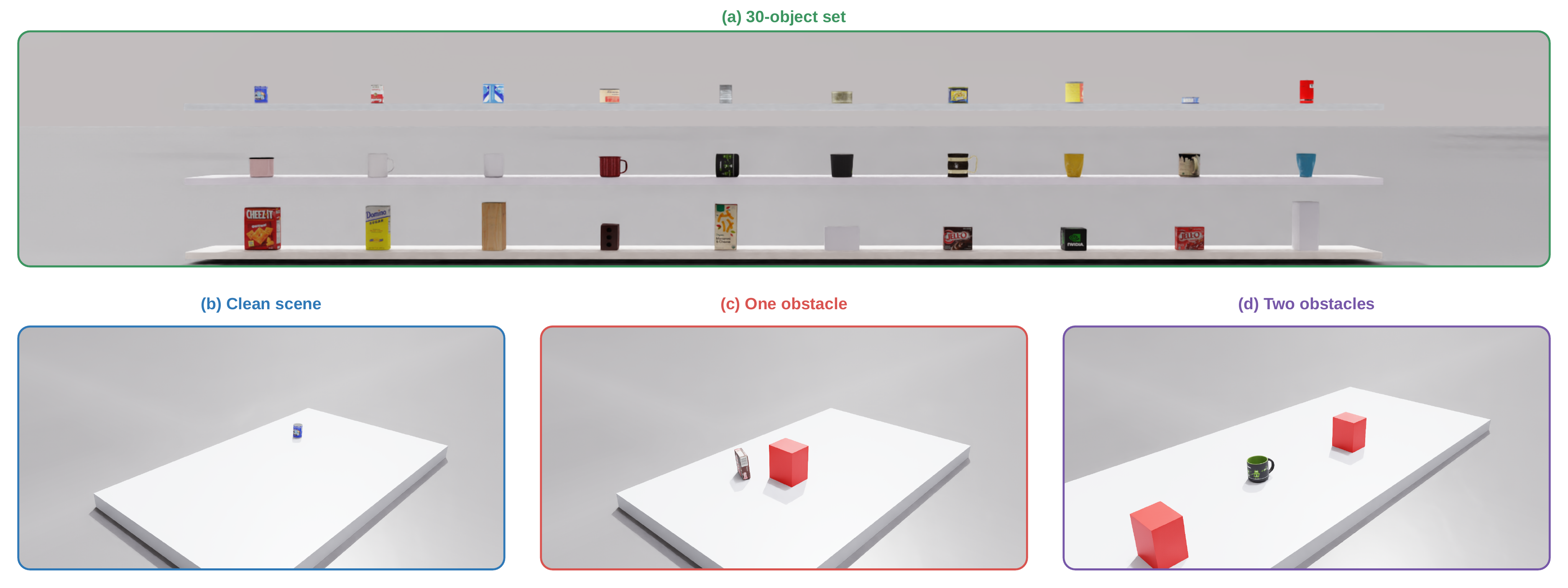}
    \caption{Execution-grounded dataset construction. Supplied grasps are paired with placement targets and evaluated through motion planning and physics execution in clean and obstacle-containing scenes.}
    \label{fig:dataset_overview}
\end{figure*}

The primary evaluation is \emph{scene-held-out}. The same object identities may appear in training, validation, and test, but complete source-qualified scene groups belong to only one split. Candidate actions from the same underlying scene therefore cannot leak across splits. The claim is generalization to unseen scene configurations, targets, obstacle layouts, and grasp--target combinations for objects represented during training. The test set is not used for model or threshold selection.

\subsection{Tasks, Metrics, and Comparison Policies}

The joint and fixed-target tasks are reported separately. One important definition is that a ranking group is \emph{covered} when its candidate pool contains at least one full-pipeline success. An uncovered group has no successful action, so no ranking policy can solve it. Candidate-pool coverage is therefore reported separately from ranking quality.

For covered groups, success-at-$K$ is
\begin{equation}
S_K
=
\frac{1}{|\mathcal{C}|}
\sum_{q\in\mathcal{C}}
\mathbb{I}
\left[
\exists\, a \in \mathrm{TopK}(q):y_{\mathrm{pipe}}(a)=1
\right],
\label{eq:success_at_k}
\end{equation}
where $\mathcal{C}$ is the set of covered groups. The reported attempt budgets are $S_1$, $S_3$, and $S_5$. Candidate-pool solvability is evaluated separately by taking the maximum predicted success probability within the group and comparing it with the binary label that at least one successful action exists.

All baselines operate on exactly the same candidate identities. The untouched test split averages 204.1 candidates per joint group and 25.5 per fixed-target group. \emph{Random} uses a deterministic random ordering. \emph{IK only} retains candidates whose requested placement endpoint admits a collision-disabled IK solution. \emph{Detached gripper collision} places only the Panda hand/finger collision geometry at the endpoint and checks overlap with the support surface and obstacles. \emph{IK + gripper collision} requires both endpoint tests. The cuMotion stack uses motion-generation methods introduced in cuRobo~\cite{curobo2023}. \emph{cuMotion full pool} runs its complete graph-based planner over every candidate and orders successful plans by total planned trajectory duration. The proposed method ranks the full pool directly with Eq.~\ref{eq:ranking_score} before any planner result is available.

Runtime is the mean decision time per ranking group, not complete rank--plan--execute latency. IK, collision, and cuMotion include the fresh pre-execution computations required by each policy. Neural GPU execution is synchronized, and inference timing uses an NVIDIA GeForce RTX 4090 Laptop GPU. The reported neural time covers inference and ranking, excluding dataset loading, host-to-device transfer, and result-file writing.

\subsection{Real-World Cross-Embodiment Transfer Protocol}
\label{sec:real_world_setup}

\paragraph{Platform and transfer setting.}
The frozen ranker is deployed on an xArm7 with ROS~2/MoveIt~\cite{moveit2012}, a wrist-mounted RealSense D435, and FoundationPose~\cite{foundationpose2024}, without xArm-specific retraining or fine-tuning. The learned model is unchanged. Frame conversion, collision modelling, safety constraints, and MoveIt integration remain robot specific.

\paragraph{Locked cases and candidates.}
The 27 locked cases include nine for a foam brick, nine for a cylindrical can, and nine for a handled mug. These are the same object identities represented in the training dataset, so the physical study tests cross-embodiment transfer rather than unseen-object generalization. Each object has three clean, three one-obstacle, and three two-obstacle scenes. Obstacles are 3D-printed $0.10\times0.08\times0.12$~m boxes. Each object has 32 fixed grasps. For the can, these grasps are evaluated over 12 symmetry-equivalent axial phases at $30^\circ$ increments. All phase--grasp combinations are ranked together, and the formal Top-5 is selected from this combined set.

\paragraph{Scene reproduction and fresh perception.}
A read-only preview is used to reproduce each locked scene. To reproduce the locked initial object position, the automatic 20~mm gate applies only to the detected object's XY start-position error. Obstacles are manually aligned to the locked visualization, and object orientation and support are checked visually. The formal run then collects a new synchronized RGB--D observation, the exact-time camera-to-base transform, and a fresh FoundationPose estimate for ranking.

\paragraph{Rank--plan--execute and outcome.}
The model ranks the full pre-execution set before planning. Only its Top-5 are considered, in order, using geometric/table screening and the required full-sequence MoveIt preflight. The first candidate that passes preflight is executed. If none passes, the case ends with no Top-5 solution. The operator records the physical outcome when execution reaches the normal prompt. A MoveIt or controller exception before that prompt also counts as a physical failure. Post-placement pose estimation was attempted when visibility permitted, but it was not used as an evaluation metric because the wrist-camera return viewpoint does not provide consistent object visibility. For transfer analysis, a candidate is predicted planning-feasible when $p_{\mathrm{plan}}\geq0.5$. A case is predicted feasible when at least one learned Top-5 candidate exceeds this threshold. It is deployment-feasible when at least one Top-5 candidate passes full-sequence preflight. Candidate-level deployment feasibility is evaluated only for candidates that receive an independent geometric/table-screening or full-preflight feasibility label. Untested candidates are excluded.

\begin{figure*}[!t]
    \centering
    \includegraphics[width=0.195\textwidth]{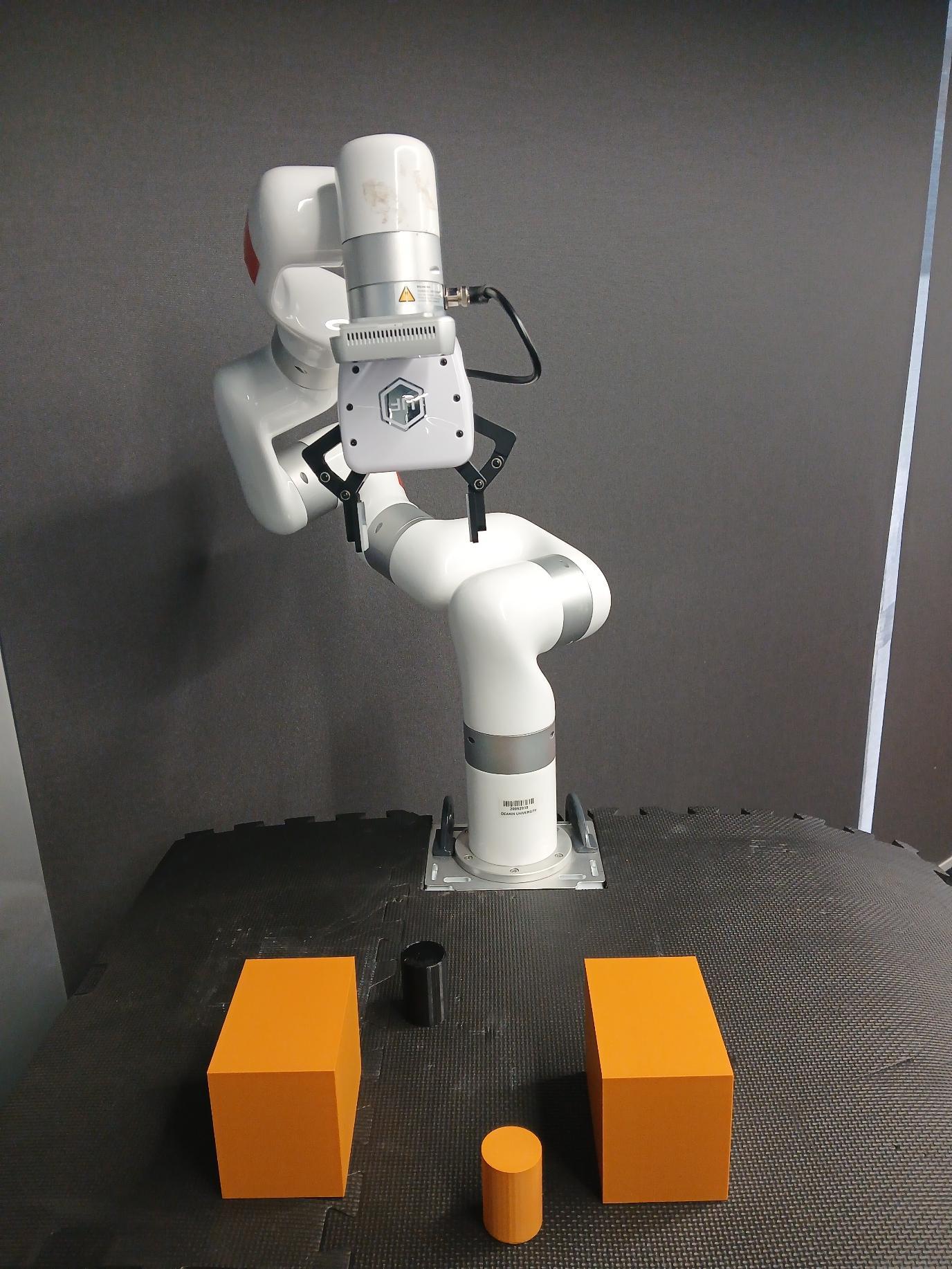}\hfill
    \includegraphics[width=0.195\textwidth]{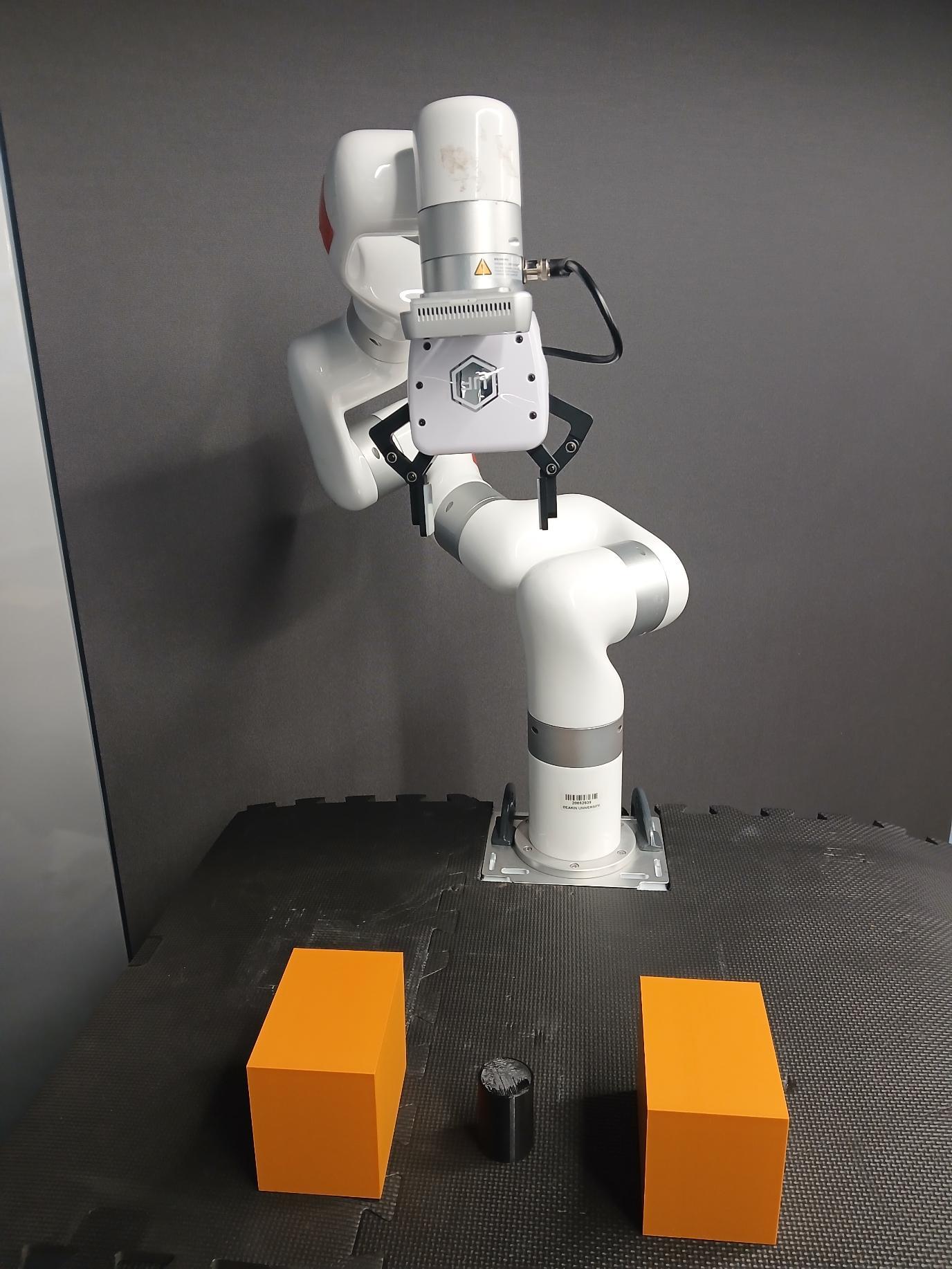}\hfill
    \includegraphics[width=0.195\textwidth]{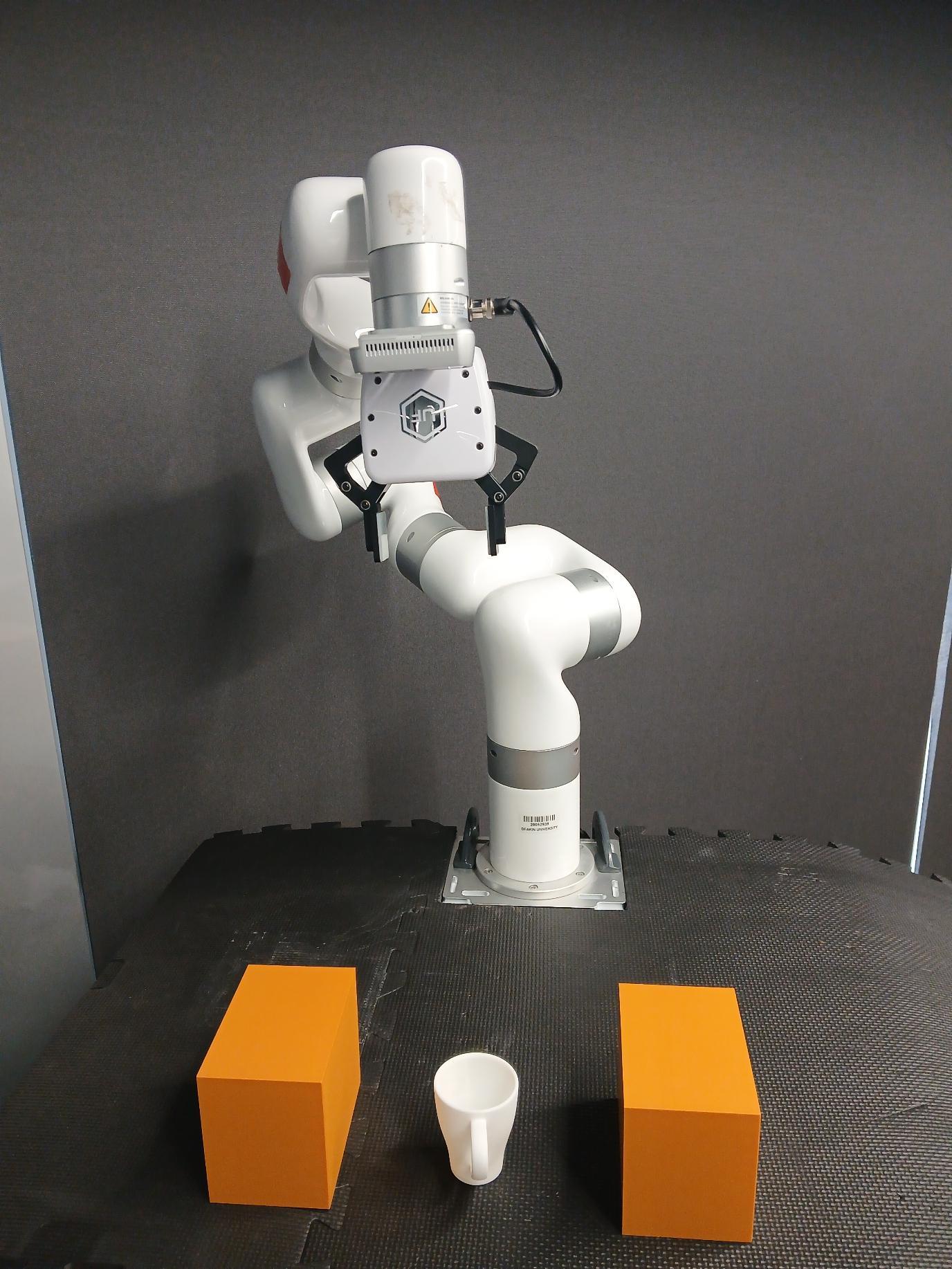}\hfill
    \includegraphics[width=0.195\textwidth]{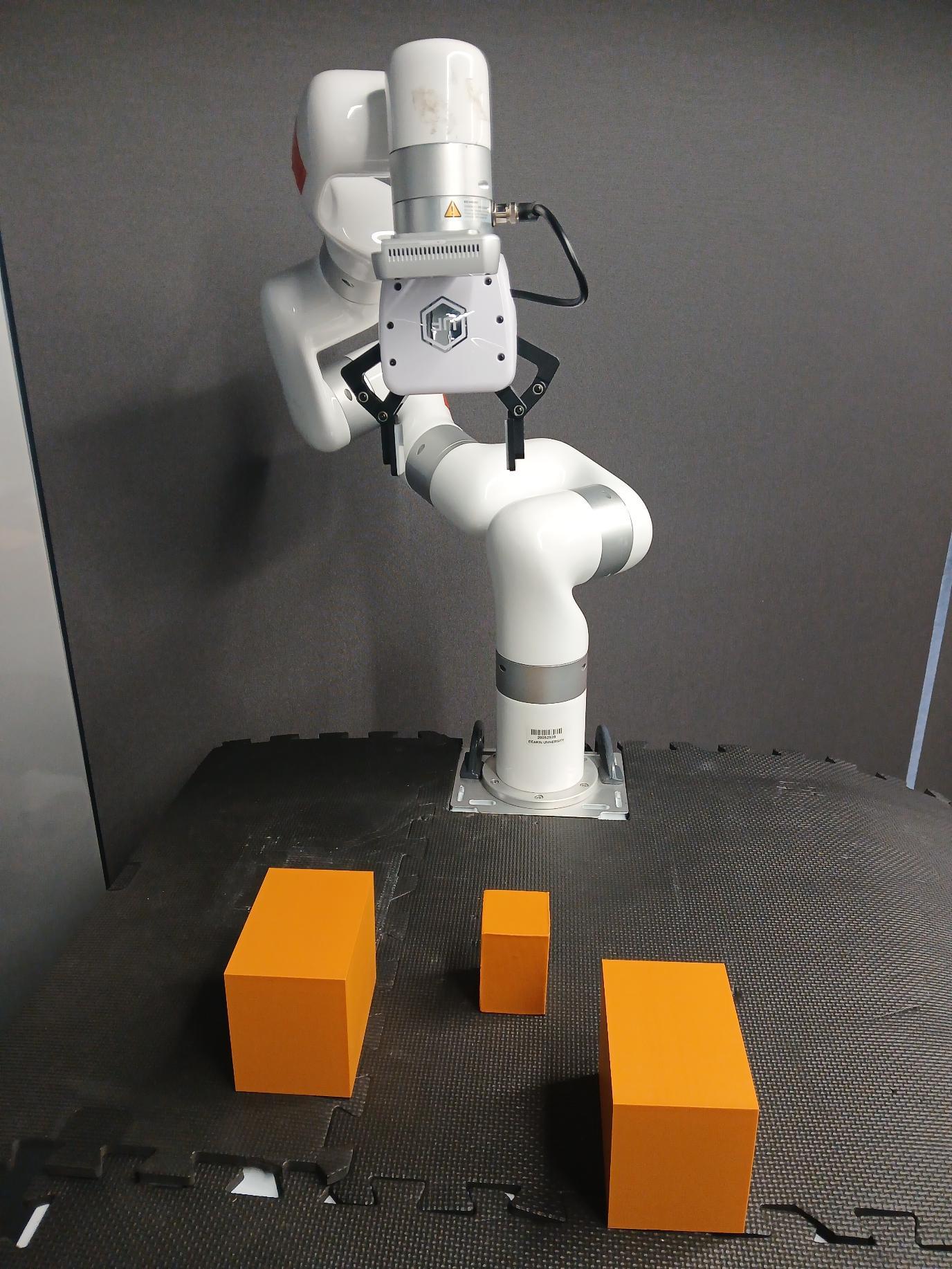}
   \caption{Real-world cross-embodiment evaluation. From left to right:
initial/target visualization for the can and representative can,
handled-mug, and foam-brick setups. The first panel shows two object
instances only for visualization; formal trials use one object.
Orange boxes are 3D-printed obstacles.}
\end{figure*}

\section{Results}
\label{sec:results}

\subsection{Ranking Performance and Runtime}

The supplied candidate pools are covered in 97.89\% of joint test groups and 50.58\% of fixed-target groups. The two settings represent different deployment scenarios: joint ranking identifies the most promising grasp--target combination when several placements are acceptable, whereas fixed-target ranking selects a grasp for a destination specified in advance. Table~\ref{tab:test_baseline_comparison} reports $S_K$ on covered groups and mean decision time for the designated frozen seed-42 checkpoint.

\begin{table*}[!t]
\centering
\caption{Frozen seed-42 ranking on the untouched test split. $S_K$ is evaluated on covered groups (97.89\% joint, 50.58\% fixed target); runtime is mean decision time per group.}
\label{tab:test_baseline_comparison}
\small
\setlength{\tabcolsep}{5pt}
\begin{tabular}{llcccc}
\toprule
Task & Method & $S_1$ (\%) & $S_3$ (\%) & $S_5$ (\%) & Time/group (s) \\
\midrule
Joint
& Random & 15.95 & 37.07 & 52.16 & $<0.001$ \\
& IK only & 20.26 & 46.98 & 60.78 & 0.254 \\
& Detached gripper collision & 39.66 & 69.40 & 82.33 & 4.990 \\
& IK + gripper collision & 41.81 & 71.55 & 84.91 & 5.243 \\
& cuMotion full pool & 72.84 & 91.38 & 94.40 & 255.746 \\
& \textbf{Proposed} & \textbf{85.78} & \textbf{97.84} & \textbf{99.14} & \textbf{0.232} \\
\midrule
Fixed target
& Random & 28.26 & 53.08 & 65.28 & $<0.001$ \\
& IK only & 33.16 & 60.48 & 72.47 & 0.032 \\
& Detached gripper collision & 53.28 & 77.16 & 82.27 & 0.624 \\
& IK + gripper collision & 54.95 & 78.42 & 82.48 & 0.655 \\
& cuMotion full pool & 59.65 & 73.31 & 74.66 & 31.968 \\
& \textbf{Proposed} & \textbf{79.67} & \textbf{94.47} & \textbf{97.60} & \textbf{0.029} \\
\bottomrule
\end{tabular}
\end{table*}

The proposed method is strongest at every $K$ in both tasks. Against full-pool cuMotion, $S_1$ improves by 12.94 percentage points in joint ranking and 20.02 points in fixed-target ranking. The fixed-target result is especially informative because all candidates share the same destination, so the ranking must distinguish how the grasp and induced object-relative pose affect the action rather than choosing an easier target. cuMotion observes planning feasibility, but $S_K$ is scored on full-pipeline execution success including downstream physics. Exhaustive planning is therefore not an oracle for final placement success. As a diagnostic, ranking by $p_{\mathrm{plan}}$ alone reaches $S_1=78.45\%$ for joint ranking and $76.02\%$ for fixed-target ranking, compared with $85.78\%$ and $79.67\%$ for the full hierarchical score, showing that conditional execution prediction contributes beyond planning-feasibility prediction.

The computational difference is also large. Full-pool cuMotion requires 255.746~s per joint group and 31.968~s per fixed-target group. The proposed ranker requires 0.232~s and 0.029~s. Across all 48,376 test candidates, synchronized forward computation takes 55.02~s, or about 1.14~ms per candidate. Using the frozen ranking with the stored planning outcomes, the first feasible plan appears after 1.17 planner evaluations on average for joint ranking and 1.47 for fixed-target ranking. The median is one call and the 90th percentile is two calls in both settings, compared with mean pool sizes of 204.1 and 25.5 candidates, respectively. Counterfactual rank-to-plan wall-clock latency is not reported because cached planning times were collected in the original exhaustive order.


\subsection{Seed Robustness}

Independent models are trained with seeds 42, 43, and 44. Table~\ref{tab:multiseed_results} reports their mean and sample standard deviation, whereas Table~\ref{tab:test_baseline_comparison} uses the designated frozen seed-42 checkpoint for the policy comparison. On the untouched test set, joint $S_1$ is $85.63\pm1.08\%$ and fixed-target $S_1$ is $79.84\pm0.16\%$. $S_5$ remains above 97\% in both tasks.

\begin{table}[!b]
\centering
\caption{Three-seed ranking performance (mean $\pm$ sample standard deviation).}
\label{tab:multiseed_results}
\small
\setlength{\tabcolsep}{3.5pt}
\resizebox{\columnwidth}{!}{%
\begin{tabular}{llcccc}
\toprule
Split & Task & Coverage & $S_1$ & $S_3$ & $S_5$ \\
\midrule
Val. & Joint & 100.00\% & $85.79\pm0.64$\% & $98.87\pm0.24$\% & $99.44\pm0.24$\% \\
Val. & Fixed target & 51.21\% & $81.05\pm0.41$\% & $93.89\pm0.12$\% & $96.88\pm0.16$\% \\
\midrule
Test & Joint & 97.89\% & $85.63\pm1.08$\% & $97.70\pm0.25$\% & $99.28\pm0.25$\% \\
Test & Fixed target & 50.58\% & $79.84\pm0.16$\% & $94.86\pm0.34$\% & $97.43\pm0.16$\% \\
\bottomrule
\end{tabular}%
}
\end{table}

\subsection{Validation Ablation Study}

Three design choices are ablated using the validation split only. The test split is
not used for ablation selection or interpretation. All variants use the same
training and deterministic evaluation protocol with seeds 42, 43, and 44.
Table~\ref{tab:ablation_results} reports covered-group ranking performance.

\begin{table}[!b]
\centering
\caption{Three-seed validation ablations (mean $\pm$ sample standard deviation); $S_K$ is evaluated on covered groups.}
\label{tab:ablation_results}
\small
\setlength{\tabcolsep}{3.5pt}
\resizebox{\columnwidth}{!}{%
\begin{tabular}{lccc}
\toprule
Variant & Joint $S_1$ & Fixed $S_1$ & Fixed $S_5$ \\
\midrule
Direct final-success head
& $83.40\pm0.88$\% & $79.13\pm0.42$\% & $96.53\pm0.12$\% \\
w/o object-relative pose
& $66.95\pm1.29$\% & $64.50\pm0.42$\% & $89.43\pm0.12$\% \\
w/o target-object hypothesis
& $\mathbf{86.78\pm0.24}$\% & $79.68\pm0.26$\% & $\mathbf{97.22\pm0.21}$\% \\
\textbf{Full hierarchical model}
& $85.79\pm0.64$\% & $\mathbf{81.05\pm0.41}$\% & $96.88\pm0.16$\% \\
\bottomrule
\end{tabular}%
}
\end{table}

\begin{table}[!b]
\centering
\caption{Solvable/unsolvable candidate-pool prediction on the untouched test split.}
\label{tab:group_feasibility}
\small
\setlength{\tabcolsep}{3.2pt}
\resizebox{\columnwidth}{!}{%
\begin{tabular}{lrrrrccc}
\toprule
Task & Groups & Solvable & Unsolvable & AUROC & Bal. Acc. & Recall$_{\mathrm{sol}}$ & Recall$_{\mathrm{unsol}}$ \\
\midrule
Joint & 237 & 232 & 5 & 99.66\% & -- & -- & -- \\
Fixed target & 1,896 & 959 & 937 & 97.77\% & 94.30\% & 94.89\% & 93.70\% \\
\bottomrule
\end{tabular}%
}
\end{table}

\begin{table*}[!t]
\centering
\caption{Real-world xArm7 results. Plan+ indicates at least one learned Top-5 candidate passing full-sequence preflight; feasibility agreement uses independent deployment labels.}
\label{tab:real_world_results}
\small
\begin{minipage}[t]{0.53\textwidth}
\centering
\textbf{(a) Physical execution outcomes}\\[2pt]
\setlength{\tabcolsep}{4.0pt}
\renewcommand{\arraystretch}{0.94}
\begin{tabular}{lrrrrr}
\toprule
Object & Cases & Plan+ & Success & Exec. fail & No plan \\
\midrule
Foam brick & 9 & 3 & 3 & 0 & 6 \\
Can        & 9 & 6 & 5 & 1 & 3 \\
Mug        & 9 & 7 & 5 & 2 & 2 \\
\midrule
\textbf{Overall} & \textbf{27} & \textbf{16} & \textbf{13} & \textbf{3} & \textbf{11} \\
\bottomrule
\end{tabular}
\end{minipage}
\hfill
\begin{minipage}[t]{0.43\textwidth}
\centering
\textbf{(b) Deployment-feasibility agreement}\\[2pt]
\setlength{\tabcolsep}{4.0pt}
\renewcommand{\arraystretch}{0.94}
\begin{tabular}{lrrrrr}
\toprule
Level & TP & TN & FP & FN & Raw agr. \\
\midrule
Case ($N=27$) & 15 & 5 & 6 & 1 & 74.07\% \\
Candidate ($N=239$) & 13 & 190 & 33 & 3 & 84.94\% \\
\bottomrule
\end{tabular}
\end{minipage}
\end{table*}

The object-relative pose has the largest effect: removing it lowers joint $S_1$
from 85.79\% to 66.95\% and fixed-target $S_1$ from 81.05\% to 64.50\%. This shows
that the object-relative pose induced by the grasp is a key cue for placement
ranking. Replacing the hierarchical heads with a single final-success head gives
a smaller but consistent top-1 drop, supporting separate planning and execution
prediction. The target-object point-cloud hypothesis is not uniformly better:
removing it slightly improves joint $S_1$ and fixed-target $S_5$, but reduces
fixed-target $S_1$ to 79.68\%. It is retained because fixed-target $S_1$ directly
measures the intended setting in which a destination is specified and the robot
must select the grasp most likely to succeed.

\subsection{Detecting Unsolvable Candidate Pools}

Ranking requires at least one successful action in the supplied pool. A group is labeled \emph{solvable} when at least one candidate succeeds and \emph{unsolvable} otherwise. Pool solvability is scored by
\begin{equation}
p_{\max}(\mathcal{A})=\max_{a_{ij}\in\mathcal{A}}p_{\mathrm{succ}}^{ij}.
\label{eq:pool_solvability_score}
\end{equation}
A low score indicates that candidate generation should be revisited rather than repeatedly planning unlikely actions. The joint test set has 232 solvable and only five unsolvable groups, for which the score gives 99.66\% AUROC. Fixed-target evaluation is balanced (959 solvable, 937 unsolvable) and reaches 97.77\% AUROC and 94.30\% balanced accuracy using a validation-selected threshold. Solvable- and unsolvable-group recall are 94.89\% and 93.70\%.


\subsection{Fresh-Execution Label Validation and Repeatability}

A label-blind study re-executes 50 fixed-target tasks selected before reading stored outcomes. Planning, final-pipeline, and three-class outcomes match the stored labels in all 50 cases (31 planning failures, one downstream failure, 18 successes). Fifteen tasks are repeated three times, giving 80 fresh executions in total, and all 45 repeats match the stored outcome. This supports simulator-label repeatability.

\subsection{Real-World Cross-Embodiment Transfer}
\label{sec:real_world_results}

Table~\ref{tab:real_world_results} summarizes the physical outcomes and deployment-feasibility results.

\paragraph{Physical execution outcomes.}
Of 27 locked cases, 16 pass Top-5 full-sequence preflight and begin physical execution. Thirteen cases complete end to end (13/27, 48.15\%), corresponding to 13/16 (81.25\%) success among executed cases. Three fail during execution and 11 have no Top-5 solution. One failure occurs during lift after successful preflight and grasp closure (MoveIt status~6), so it is a downstream execution failure rather than a planning-feasibility error.

\paragraph{Deployment-feasibility transfer.}
Case-level $p_{\mathrm{plan}}$ gives TP=15, TN=5, FP=6, and FN=1, or 20/27 (74.07\%) raw agreement. Across 239 candidate-level deployment-feasibility labels, TP=13, TN=190, FP=33, and FN=3, giving 81.25\% recall, 85.20\% specificity, 83.23\% balanced accuracy, 28.26\% precision, and 84.94\% raw agreement. The low precision reflects optimistic predictions under xArm7-specific constraints. Labels combine geometric/table screening with applicable full-sequence checks and do not represent 239 MoveIt calls.

\paragraph{Remaining errors.}
The six case-level false positives do not all have the same cause. They include robot-specific feasibility errors, Top-5 ranking misses, and can cases where axial symmetry changes the robot configuration. Sec.~\ref{sec:discussion} discusses these cases in more detail. 

\section{Discussion}
\label{sec:discussion}

\subsection{Planning Feasibility, Execution, and Fixed-Target Ranking}

Full-pool cuMotion is stronger than endpoint-only checks at $S_1$,
confirming that path feasibility matters. The learned ranker improves
further because plan existence alone does not determine full placement
success. Ablations show that the object-relative pose is the strongest
cue, while hierarchical planning/execution supervision gives an
additional top-1 gain. Fixed-target ranking isolates grasp--placement
coupling because all candidates share the same destination.

\subsection{Candidate-Pool Coverage and System Role}
The ranker works with rather than replaces the motion planner. The 50.58\% fixed-target coverage shows that many specified targets contain no successful action, which is a candidate-pool limitation rather than a ranking error. Solvability prediction can trigger candidate regeneration, while the $S_1$--$S_5$ gap shows that later-ranked candidates can recover when the first choice is rejected.

\subsection{Cross-Embodiment Transfer and Failure Modes}
The frozen ranker transfers to a different robot and planning stack without xArm-specific retraining, although frame conversion, collision modelling, safety constraints, and MoveIt integration remain robot specific. The five case-level and 190 candidate-level true negatives show that useful rejection information transfers across embodiments.

The six case-level false positives have different causes. Some are genuine robot-specific feasibility errors, while others are Top-5 ranking misses. Exhaustive audits find feasible candidates below Top-5 in one one-obstacle brick scene (about rank~9) and two one-obstacle can scenes (about ranks~85 and~18). For the can, rotations about its symmetry axis leave the desired object placement approximately unchanged but change the gripper/TCP orientation, so object-equivalent poses can have different xArm7 kinematic feasibility. The single case-level false negative is a one-obstacle mug scene that plans and executes successfully despite low predicted feasibility. A separate mug trial passes preflight but fails during lift after grasp closure, illustrating why $p_{\mathrm{plan}}$ and complete task success must remain separate. Future ranking should account more directly for robot kinematics and object symmetry.

\section{Conclusion}

This paper presented execution-aware pre-execution ranking for
grasp-conditioned robotic placement. Using only information available
before motion planning, the model jointly estimates placement-planning
feasibility and execution success conditioned on a valid plan, allowing
supplied grasp--placement candidates to be ranked before expensive
planning is invoked.

On the 30-object scene-held-out benchmark, three-seed $S_1$ reaches
$85.63 \pm 1.08\%$ for joint ranking and $79.84 \pm 0.16\%$ for
fixed-target ranking. The frozen ranking places the first feasible plan
within 1.17 planner evaluations on average for joint ranking and 1.47
for fixed-target ranking, showing how learned ranking can complement
rather than replace motion planning. Fresh execution reproduces all
50 stored simulator outcomes. Under frozen transfer to xArm7/MoveIt,
13/27 locked cases complete end to end, while 13/16 executed cases
succeed. Candidate-level deployment-feasibility prediction reaches
81.25\% recall, 85.20\% specificity, and 83.23\% balanced accuracy.

Together, these results show that execution-aware ranking can move
promising actions earlier in the planning queue while retaining a
clear separation between learned prediction, motion planning, and
physical execution. The remaining limitations are candidate-pool
coverage, embodiment-specific feasibility differences, and
symmetry-sensitive ranking. Future work should incorporate more
explicit embodiment-aware information, improve candidate regeneration,
and evaluate broader object generalization.

\section*{AI Use Disclosure}
OpenAI ChatGPT and Anthropic Claude assisted with language editing,
manuscript restructuring, and drafting portions of the text. All
technical content and results were verified by the authors.
\bibliographystyle{IEEEtran}
\bibliography{IEEEfull}

@inproceedings{r2,
  title={Where to relocate?: Object rearrangement inside cluttered and confined environments for robotic manipulation},
  author={Cheong, Sang Hun and Cho, Brian Y and Lee, Jinhwi and Kim, ChangHwan and Nam, Changjoo},
  booktitle={IEEE International Conference on Robotics and Automation (ICRA)},
  year={2020}
}

@article{extra,
  title={Validating an object placement planner for robotic pick-and-place tasks},
  author={Harada, Kensuke and Tsuji, Tokuo and Nagata, Kazuyuki and Yamanobe, Natsuki and Onda, Hiromu},
  journal={Robotics and Autonomous Systems},
  year={2014},

}

@article{r3,
  title={Learning to place unseen objects stably using a large-scale simulation},
  author={Noh, Sangjun and Kang, Raeyoung and Kim, Taewon and Back, Seunghyeok and Bak, Seongho and Lee, Kyoobin},
  journal={IEEE Robotics and Automation Letters},
  year={2024}
}

@inproceedings{r4,
  title={Placing by Touching: An empirical study on the importance of tactile sensing for precise object placing},
  author={Lach, Luca and Funk, Niklas and Haschke, Robert and Lemaignan, S{\'e}verin and Ritter, Helge Joachim and Peters, Jan and Chalvatzaki, Georgia},
  booktitle={IEEE/RSJ International Conference on Intelligent Robots and Systems (IROS)},
  year={2023}
}

@article{r5,
  title={Pick and place planning is better than pick planning then place planning},
  author={Shanthi, Mohanraj Devendran and Hermans, Tucker},
  journal={IEEE Robotics and Automation Letters},
  year={2024},

}

@inproceedings{r6,
  title={B4P: Simultaneous grasp and motion planning for object placement via parallelized bidirectional forests and path repair},
  author={Leebron, Benjamin H and Ren, Kejia and Chen, Yiting and Hang, Kaiyu},
  booktitle={IEEE/RSJ International Conference on Intelligent Robots and Systems (IROS)},
  year={2025}
}

@article{r7,
  title={Robotic pick-and-place with uncertain object instance segmentation and shape completion},
  author={Gualtieri, Marcus and Platt, Robert},
  journal={IEEE robotics and automation letters},
  year={2021},
}

@article{r8,
  title={Learning to regrasp by learning to place},
  author={Cheng, Shuo and Mo, Kaichun and Shao, Lin},
  journal={arXiv preprint arXiv:2109.08817},
  year={2021}
}

@article{r9,
  title={Optimizing Robotic Placement via Grasp-Dependent Feasibility Prediction},
  author={Liu, Tianyuan and Dazeley, Richard and Champion, Benjamin and Cosgun, Akan},
  journal={Australasian Conference on Robotics and Automation (ACRA)},
  year={2025}
}

@article{r10,
  title={Learning feasibility for task and motion planning in tabletop environments},
  author={Wells, Andrew M and Dantam, Neil T and Shrivastava, Anshumali and Kavraki, Lydia E},
  journal={IEEE robotics and automation letters},
  year={2019}
}

@inproceedings{r11,
  title={Learning models as functionals of signed-distance fields for manipulation planning},
  author={Driess, Danny and Ha, Jung-Su and Toussaint, Marc and Tedrake, Russ},
  booktitle={Conference on Robot Learning},
  year={2022}
}

@inproceedings{r13,
  title={On the continuity of rotation representations in neural networks},
  author={Zhou, Yi and Barnes, Connelly and Lu, Jingwan and Yang, Jimei and Li, Hao},
  booktitle={IEEE/CVF conference on computer vision and pattern recognition},
  year={2019}
}

@article{r14,
  title={Fourier features let networks learn high frequency functions in low dimensional domains},
  author={Tancik, Matthew and Srinivasan, Pratul and Mildenhall, Ben and Fridovich-Keil, Sara and Raghavan, Nithin and Singhal, Utkarsh and Ramamoorthi, Ravi and Barron, Jonathan and Ng, Ren},
  journal={Advances in neural information processing systems},
  year={2020}
}

@article{mahler2018dex,
  title={Dex-Net 2.0: Deep Learning to Plan Robust Grasps with Synthetic Point Clouds and Analytic Grasp Metrics},
  author={Mahler, Jeffrey and Liang, Jacky and Niyaz, Sherdil and Aubry, Mathieu and Laskey, Michael and Doan, Richard and Liu, Xinyu and Ojea, Juan Aparicio and Goldberg, Ken},
  year={2018}
}

@inproceedings{moreau2023imposing,
  title={Imposing: Implicit pose encoding for efficient visual localization},
  author={Moreau, Arthur and Gilles, Thomas and Piasco, Nathan and Tsishkou, Dzmitry and Stanciulescu, Bogdan and de La Fortelle, Arnaud},
  booktitle={IEEE/CVF Winter Conference on Applications of Computer Vision},
  year={2023}
}

@article{kuang2026dex4d,
  title={Dex4D: Task-Agnostic Point Track Policy for Sim-to-Real Dexterous Manipulation},
  author={Kuang, Yuxuan and Park, Sungjae and Fragkiadaki, Katerina and Tulsiani, Shubham},
  journal={arXiv preprint arXiv:2602.15828},
  year={2026}
}

@inproceedings{deitke2023objaverse,
  title={Objaverse: A Universe of Annotated 3D Objects},
  author={Deitke, Matt and Schwenk, Dustin and Kembhavi, Aniruddha and Farhadi, Ali and Kolve, Eric and Mottaghi, Roozbeh},
  booktitle={IEEE/CVF Conference on Computer Vision and Pattern Recognition (CVPR)},
  year={2023}
}

@misc{isaacsim,
  title = {NVIDIA Isaac Sim},
  howpublished = {\url{https://github.com/isaac-sim/IsaacSim}},
  note = {Accessed: Sep. 14, 2026}
}

@inproceedings{pointnetpp2017,
  title={PointNet++: Deep Hierarchical Feature Learning on Point Sets in a Metric Space},
  author={Qi, Charles Ruizhongtai and Yi, Li and Su, Hao and Guibas, Leonidas J.},
  booktitle={Advances in Neural Information Processing Systems},
  volume={30},
  pages={5099--5108},
  year={2017}
}

@inproceedings{foundationpose2024,
  title={FoundationPose: Unified 6D Pose Estimation and Tracking of Novel Objects},
  author={Wen, Bowen and Yang, Wei and Kautz, Jan and Birchfield, Stan},
  booktitle={IEEE/CVF Conference on Computer Vision and Pattern Recognition (CVPR)},
  pages={17868--17879},
  year={2024}
}

@article{moveit2012,
  title={MoveIt!},
  author={Chitta, Sachin and Sucan, Ioan Alexandru and Cousins, Steve},
  journal={IEEE Robotics \& Automation Magazine},
  volume={19},
  number={1},
  pages={18--19},
  year={2012},
  doi={10.1109/MRA.2011.2181749}
}

@inproceedings{curobo2023,
  title={CuRobo: Parallelized Collision-Free Robot Motion Generation},
  author={Sundaralingam, Balakumar and Hari, Siva Kumar Sastry and Fishman, Adam and Garrett, Caelan and Van Wyk, Karl and Blukis, Valts and Millane, Alexander and Oleynikova, Helen and Handa, Ankur and Ramos, Fabio and Ratliff, Nathan and Fox, Dieter},
  booktitle={IEEE International Conference on Robotics and Automation (ICRA)},
  pages={8112--8119},
  year={2023},
  doi={10.1109/ICRA48891.2023.10160765}
}

@inproceedings{aitbouhsain2023action,
  title={Learning to Predict Action Feasibility for Task and Motion Planning in 3D Environments},
  author={Ait Bouhsain, Smail and Alami, Rachid and Sim{\'e}on, Thierry},
  booktitle={IEEE International Conference on Robotics and Automation (ICRA)},
  pages={3736--3742},
  year={2023},
  doi={10.1109/ICRA48891.2023.10161114}
}

@inproceedings{aitbouhsain2023multitask,
  title={Simultaneous Action and Grasp Feasibility Prediction for Task and Motion Planning Through Multi-Task Learning},
  author={Ait Bouhsain, Smail and Alami, Rachid and Sim{\'e}on, Thierry},
  booktitle={IEEE/RSJ International Conference on Intelligent Robots and Systems (IROS)},
  year={2023},
  doi={10.1109/IROS55552.2023.10341257}
}

\end{document}